\pdfoutput=1
\documentclass[11pt]{article}
\usepackage[preprint]{acl}

\usepackage{booktabs}
\usepackage{times}
\usepackage{latexsym}
\usepackage[T1]{fontenc}
\usepackage[utf8]{inputenc}
\usepackage{microtype}
\usepackage{inconsolata}
\usepackage{graphicx}
\usepackage{multirow}
\usepackage{xspace}
\usepackage{tcolorbox}
\usepackage{amsthm}

\newcommand{\disamkblogo}{%
  \raisebox{0ex}{\includegraphics[height=1.4ex]{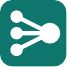}}%
}

\newcommand{\entity}[2]{%
  \href{https://gptkb.org/entity/#1}{%
    \mbox{\disamkblogo\,\texttt{#2}}%
  }%
}

\newcommand{\predicate}[2]{%
  \href{https://gptkb.org/prop/#1}{%
    \mbox{\disamkblogo\,\texttt{#2}}%
  }%
}

\newtcolorbox{mybox}[1]{colback=cyan!5!white,colframe=cyan!75!black,fonttitle=\bfseries,title=#1}

\newcommand{\ourmethod}{\textsc{GPTKB 2.0}\xspace}
\newcommand{\ourkb}{\textsc{GPTKB 2.0}\xspace}

\title{Constructing Disambiguated Knowledge Bases \\ from Large Language Models at Scale}

\author{
Yujia Hu$^{1}$
\quad
Tuan-Phong Nguyen$^2$
\quad
Simon Razniewski$^1$
\\ 
\\
$^1$ScaDS.AI Dresden/Leipzig \& TU Dresden, Germany
\\
$^2$ VNU University of Engineering and Technology, Hanoi, Vietnam
\\
{\small\texttt{\{yujia.hu,simon.razniewski\}@tu-dresden.de \quad tuanphong@vnu.edu.vn }}
}

\begin{document}
\maketitle
\begin{abstract}
Automated Knowledge Base Construction (AKBC) is a core NLP task, and recent work proposes generating knowledge bases directly from large language models (LLMs), treating the model itself as the knowledge source. However, LLMs natively possess no representation of entities, leading to duplicate entries as well as conflations. 

We propose \ourmethod, a methodology for constructing \textit{disambiguated KBs directly from LLMs}. \ourmethod incorporates on-the-fly disambiguation of entities, relations and classes, and is meticulously designed to satisfy both scalability and disambiguation accuracy. We analyze the central design decisions and characterize the trade-offs between accuracy, scale, and cost. We execute \ourmethod at scale, obtaining a materialized KB containing over 1M disambiguated entities and 38.4M triples. This represents the first million-scale LLM-native KB with explicit internal canonicalization of entities, relations, and classes, a significant departure from prior Wikimedia-centric works. 
\end{abstract}

\section{Introduction}

\begin{figure}[t]
    \centering
    \includegraphics[
    width=\linewidth,
    trim=17cm 2.5cm 17cm 2.5cm,
    clip
]{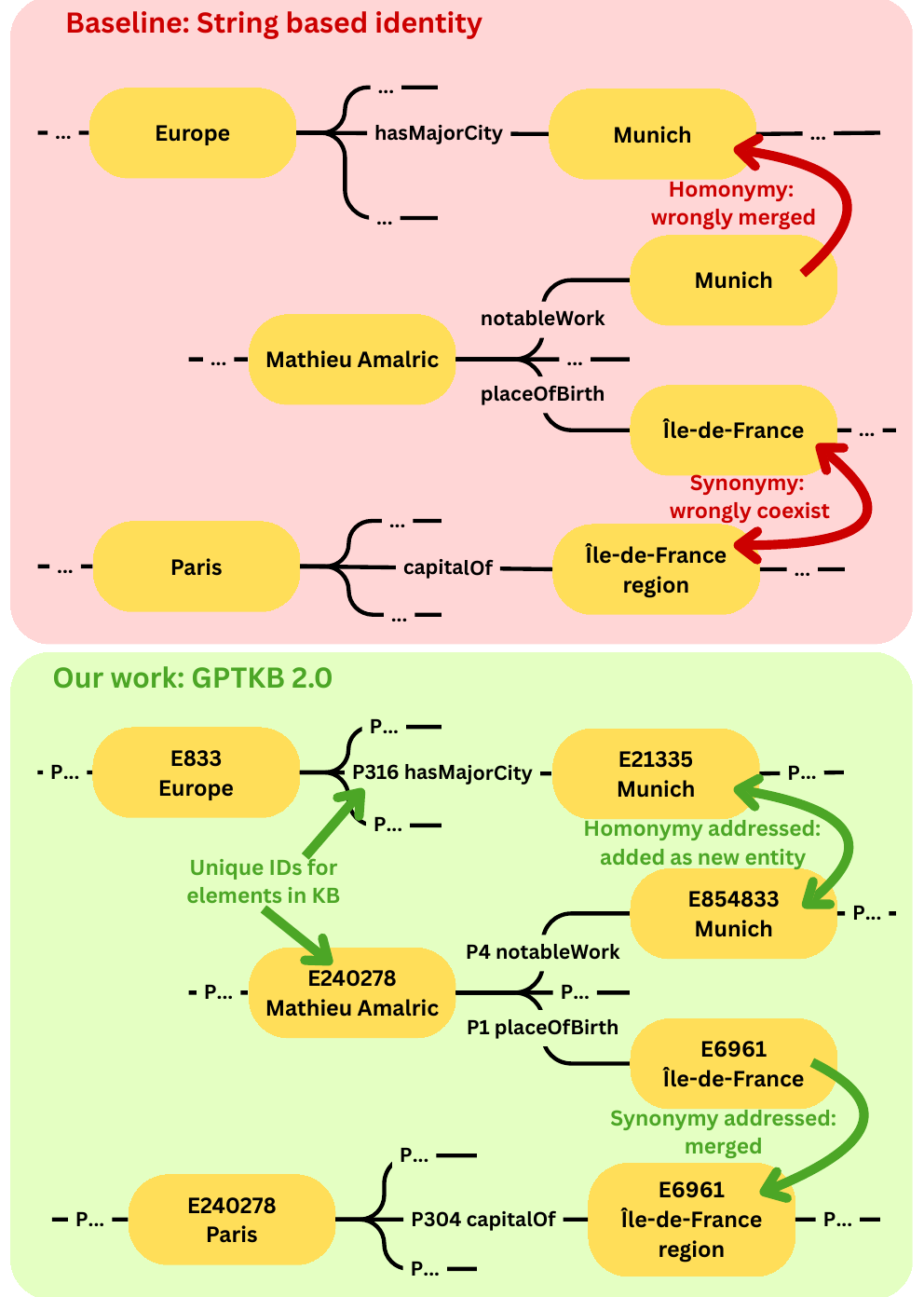}
    \caption{The challenges of syonymy and homonymy that our \ourmethod addresses.}
    \label{fig:teaser}
\end{figure}

\paragraph{Motivation and Problem.}
Automated knowledge base construction (AKBC) has long been a central vision of NLP and AI \cite{Brin1998DIPRE,Agichtein2000Snowball,Yates2007TextRunner}, with entity disambiguation recognized as a core challenge from the outset. Since 2007, Wikipedia and later Wikidata have become the de-facto backbone for open-domain entity canonicalization \cite{yago,dbpedia}, with comparatively few approaches attempting to move beyond these resources.

More recently, large language models (LLMs) have emerged as implicit repositories of factual knowledge~\cite{petroni-etal-2019-language}, inspiring a line of work that constructs knowledge bases directly from LLMs by materializing their parametric knowledge in structured form~\cite{mango, cohen-etal-2023-crawling, gptkbv1, gptkbv1.5,parovic-etal-2025-generating, giordano2026beyond}. However, these approaches largely rely on surface strings as entity identifiers, which leads to two complementary failure modes. As illustrated in Figure~\ref{fig:teaser} (top), string-based canonicalization incorrectly merges homonymous entities whose labels coincide (e.g., conflating the city \entity{E21335}{Munich} and the film \entity{E854833}{Munich}), while failing to merge synonymous entities whose labels differ (e.g., \textit{Île-de-France} and \textit{Île-de-France region}). The same issue applies analogously to relations and classes. The underlying problem is structural: LLMs are trained on surface strings, without stable identifiers for the entities, relations, and classes those strings denote. Surface strings alone are insufficient both to distinguish homonymous entities and to recognize synonymous ones.



\paragraph{Approach and Contribution.}
To address this limitation, we propose \ourkb, a method for scalable construction of disambiguated knowledge bases directly from an LLM, through refined knowledge representation, context-guided disambiguation, and scaling via cautious parallelization. Starting from a seed entity, the pipeline iteratively expands the KB by alternating elicitation and consolidation along the subject--object frontier. Our key observation is that the eliciting triple itself, together with descriptions of existing entities, typically provides sufficient context for disambiguation. Following Wikidata, we represent each entity by a unique identifier paired with a label and a generated textual description. For example, in the triple \textit{[Mathieu Amalric, notableWork, Munich]} shown in Figure~\ref{fig:teaser}, the description attached to the existing \entity{E21335}{Munich} entity identifies it as a city, thereby allowing the newly generated mention to be identified as novel entity (film) instead.

The main contributions of this work are:
\begin{enumerate}
\item We propose a parallelized, context-guided pipeline for KB construction that uses an LLM as the knowledge source. Starting from a seed entity, the pipeline expands the KB by iterating elicitation and consolidation along the subject-object frontier, addressing synonymy and homonymy through on-the-fly disambiguation. Our code is publicly available.\footnote{\url{https://github.com/Knowledge-aware-AI/GPTKB2_disambiguation_on_the_fly}}
\item We construct GPTKB 2.0, a KB of 38.4M triples and 1.6M entities that resolves synonymy and homonymy at scale, consolidating 2,316,275 surface labels into 1,592,185 canonical entities and correctly distinguishing 131,990 homonymous entities that share 41,275 surface labels.\footnote{\url{https://gptkb.org/}}
\item GPTKB 2.0 is, to our knowledge, the first automatically constructed large-scale disambiguated general-domain KB independent of Wikimedia sources. 36.8\% of its canonical entities are not present in Wikidata, with divergence concentrated in the long tail, demonstrating substantial coverage beyond curated resources.
\end{enumerate}

\section{Background}

\paragraph{Knowledge Base Construction}
Knowledge bases have traditionally been built from curated or textual sources. Wiki-based systems such as DBpedia~\cite{dbpedia}, YAGO~\cite{yago}, and Wikidata~\cite{wikidata} inherit much of their entity identity from human-maintained resources, so canonicalization is largely resolved by construction. Text-based systems instead extract facts directly from corpora, as in NELL~\cite{Carlson2010NELL} and OpenIE pipelines such as ReVerb~\cite{Fader2011ReVerb}; these offer broader coverage and open-domain flexibility, but often set entity consolidation aside. More recently, a third line of work has aimed to construct KBs directly from LLMs, either by recursively eliciting parametric knowledge~\cite{cohen-etal-2023-crawling, gptkbv1, gptkbv1.5} or by list-based polling ~\cite{He2023BertNet,parovic-etal-2025-generating}. This offers the chance to inspect LLM knowledge via human-comprehensible symbolic approximations, for which the term \textit{epistemic twins} was coined \cite{razniewski2026epistemic}. However, LLM-native KB construction encounters novel challenges as this removes the external structural signals that earlier systems relied on to maintain coherent entity identity.

\paragraph{Disambiguation via External Identifiers}
A standard way to manage ambiguity is to ground mentions in an external inventory such as Wikipedia or Wikidata \cite{wikidata}. In this design, the system links each mention back to a pre-existing identifier and thereby inherits that resource's canonicalization decisions~\cite{hoffart-etal-2011-robust, logeswaran-etal-2019-zero}. Wikipedia/data grounding has become the standard approach since DBpedia \cite{dbpedia} and YAGO \cite{yago}, however, it naturally limits the possible ambition and coverage of KBC projects.

\paragraph{String-Based Identity and Clustering}
At the other extreme, one can simply treat surface strings as identities. This is scalable and easy to parallelize, but ignores the fact that natural language is both synonymous and homonymous. Surface-form consolidation via clustering has a long history in KB construction~\cite{machine-knowledge, 10.1145/2661829.2662073, 10.1145/3178876.3186030, zhang-soh-2024-extract, saeed2026llmpedia}, but addresses only one side of the problem: clustering can merge synonymous labels, yet by construction cannot split a single label that denotes distinct entities.

\paragraph{Real Disambiguation with Homonymy}
Disambiguating homonyms is a much harder problem, and mostly only tackled in the lexical domain \cite{agirre1996word,navigli2009word}. Notable exceptions are attempts in \cite{moro-etal-2014-entity,Hoffart2014EmergingEntities,Hoffart2016KnowledgeAwakens}, yet none of them operating at scale. Homonymy cannot be deferred to post-hoc consolidation like synonymy, and on-the-fly disambiguation induces globally coupled decisions: each decision changes the candidate space for all others, making the process difficult to parallelize. Existing on-the-fly methods are therefore extremely limited, e.g., EDC~\cite{zhang-soh-2024-extract} (which does not address homonymy) is evaluated only on 1000 samples, and naively scaling it to 38M disambiguation decisions would require a runtime of one year. This leaves open the question of how to retain the benefits of true disambiguation without sacrificing the throughput needed for LLM-native KB construction at scale.

\paragraph{Problem Statement}
Our goal is to combine the openness of LLM-native KB construction with the rigor of explicit canonicalization, but without relying on external identifiers. This makes the task harder than entity linking, because there is no reference inventory. The challenge is therefore to assign stable internal identities while the KB is being generated, in a search space that expands dynamically as new entities and relations are elicited. We study the following question:

\begin{mybox}{Research Question}
Given one or several seed entities, how can we scalably construct a canonicalized open-domain knowledge base directly from a large language model, without using external identifiers?
\end{mybox}

\begin{table*}[t]
\centering
\small
\begin{tabular}{l l l l}
\toprule
ID & label & description & aliases \\
\midrule
E6961 & Île-de-France region & The most populous region of France... & Île-de-France \\
E21335 & Munich & The capital of the German state of Bavaria... & München \\
E854833 & Munich & A 2005 historical drama thriller film... & - \\
\bottomrule
\end{tabular}
\caption{\ourkb entity data model example.}
\label{tab:datamodel}
\end{table*}

\section{\ourmethod Approach}

\subsection{Data Model} 
\newtheorem{definition}{Definition}

Our data model represents entities, relations and classes as first-class elements with unique identifiers, enabling disambiguation across homonymous and synonymous mentions. Inspired by curated KBs like Wikidata~\cite{wikidata}, we additionally attach a textual description to each element to support context-based disambiguation.

\begin{definition}[Disambiguated Knowledge Base]
A disambiguated knowledge base is a quadruple $(T,E,R,C)$, containing a set $T$ of triples, $E$ of entities, $R$ of relations, and $C$ of classes. Hereby:
\begin{itemize}
    \item Each $t \in T$ is an 6-tuple containing a subject ID, relation ID, object ID (entity ID, class ID, or NULL for literals), and the labels of the subject, relation, and object label.
    \item Each $e\in E$ is a quadruple containing entity ID, label, description, and a set of aliases;
    \item Each $r\in R$ is a quadruple containing relation ID, label, description, and a set of aliases;
    \item Each $c\in C$ is a quadruple containing class ID, label, description, and a set of aliases;
\end{itemize}
\end{definition}
\noindent Elements are uniquely identified by their IDs rather than their labels. This allows multiple distinct entities to share the same label (e.g., several entities named \textit{Munich}), accommodating homonymy, while synonymy is captured by attaching multiple aliases to a single element. To ensure that all elements remain distinguishable, any two elements must differ in either their label or their description. Labels are stored alongside IDs in each triple for self-contained interpretability. Examples for entities are shown in Table~\ref{tab:datamodel}.

\begin{figure*}[t]
    \centering
    \includegraphics[width=1\linewidth,
                 ]{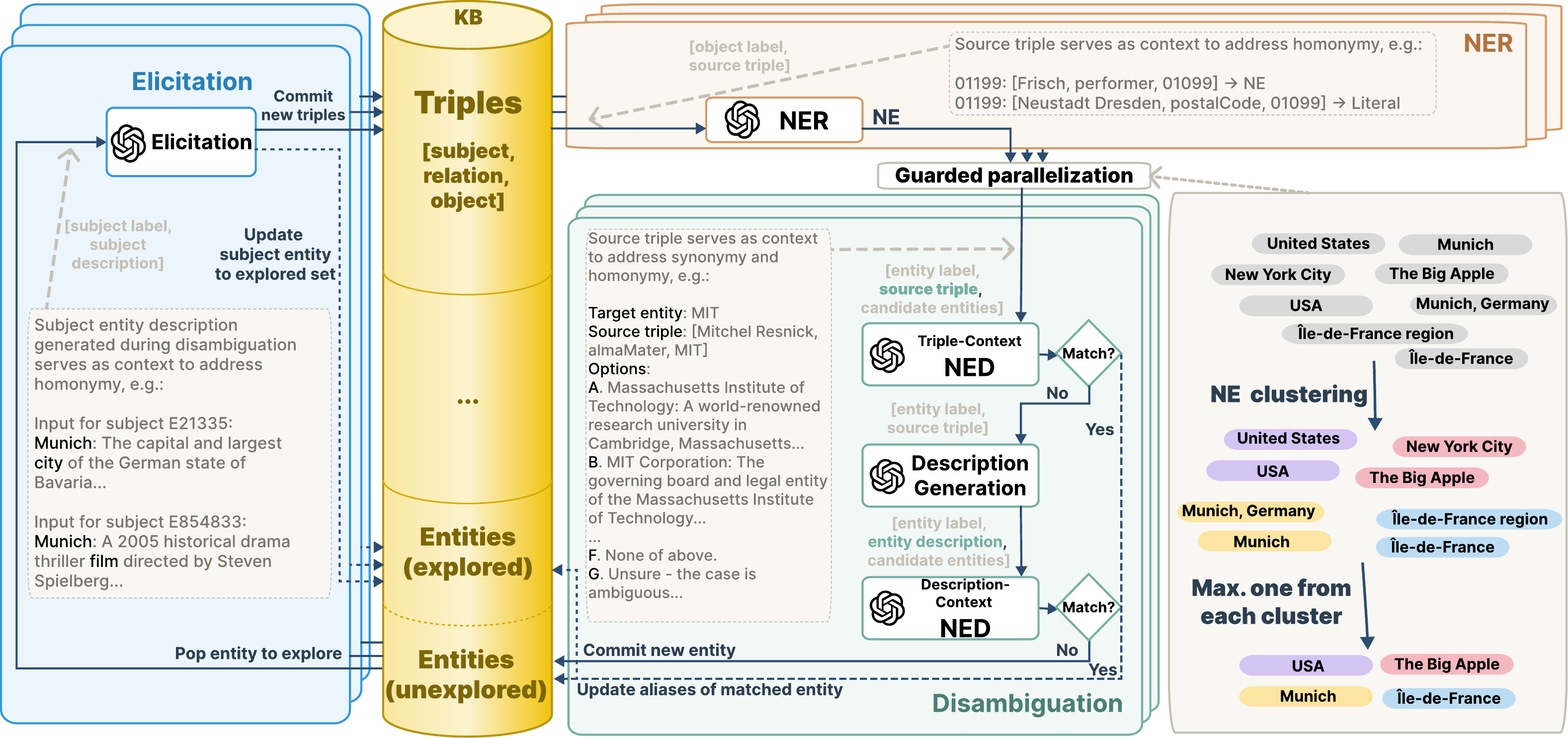}
    \caption{The \ourkb construction pipeline. Elicitation and NER are natively parallel, while NE clustering and two-step NED are used to enable guarded NED parallelization.}
    \label{fig:architecture}
\end{figure*}

\subsection{Phases}
Our construction pipeline (illustrated in Fig.~\ref{fig:architecture}) follows a recursive, context-guided paradigm characterized by on-the-fly consolidation. Starting from a seed entity, the pipeline iteratively expands the KB by alternating elicitation and consolidation along the subject--object frontier, progressing through three core operations: elicitation, named entity recognition (NER), and disambiguation. 

\subsubsection{Elicitation}
Using the prompt shown in Fig.~\ref{fig:prompt-elicitation}, the LLM elicits triples for each entity. A prompt asking simply for facts about \textit{Munich}, however, cannot on its own distinguish between the city and the 2005 Spielberg film. We therefore make elicitation context-guided: alongside the entity label, we include the entity's description in the prompt, ensuring that the elicited facts pertain to the intended entity rather than a homonymous one. Additionally, we instruct the LLM to generate at least one \textit{instanceOf} triple, where the object captures the entity's type; these class-defining triples are then routed through our class disambiguation pipeline.

\subsubsection{Named Entity Recognition}
Using the prompt shown in Fig.~\ref{fig:prompt-ner}, the LLM classifies each elicited object as either a literal or a NE. A label like \textit{01099}, however, cannot be classified on its own: it is a literal in \textit{[Neustadt Dresden, postalCode, 01099]} but a named entity (NE) in \textit{[Frisch, partOfBand, 01099]}. We therefore make NER also context-guided: alongside the object label, we include the source triple in the prompt, allowing the model to interpret the object in the relational context in which it was elicited.

\subsubsection{Disambiguation}
Disambiguation is the core operation for resolving synonymy and homonymy. Specifically, it ensures that homonymous entities are properly distinguished and accurately elicited in subsequent step, while simultaneously eliminating the redundant synonymous entities \cite{Ding2024EntGPTEL}.
NED is a challenging task regarding the performance of both candidate retrieval and disambiguation decision, which both improve with context that introduces cost-performance tradeoffs. We therefore split the process into two rounds of NED, separated by an intermediate description generation step. An ablation comparing this design to single-round NED is provided in Appendix~\ref{app:ablationonned}.
\begin{enumerate}
\item \textbf{Triple-Context NED:} For each triple object recognized as NE by NER, the $k$ most similar candidates in the KB are retrieved by label embedding similarity. As shown in Fig.~\ref{fig:architecture}, the LLM is prompted with the target entity's label, its source triple as context, and the labels and descriptions of the retrieved candidates. The model must either match the target to one of the candidates, select \textit{None of the above} (indicating a new entity), or select \textit{Unsure} (giving a NULL-ID). This round resolves the majority of both synonymy and homonymy cases: label-similarity retrieval surfaces candidates that the source triple and candidate descriptions then allow the model to confirm or reject. If a match is identified, the target is linked to the chosen candidate and its label is added to the candidate's alias set, so that future lookups can find the entity by either name. Prompt is given in Fig.~\ref{fig:prompt-ned1}.
\item \textbf{Description Generation:} Entities marked \textit{None of the above} are routed to description generation. The LLM is prompted with the new entity's label and source triple (Fig.~\ref{fig:prompt-nedg}) and produces a brief textual description. The source triple serves as disambiguating context: the description generated for \textit{Munich} in \textit{[Mathieu Amalric, notableWork, Munich]} is the film, while the description in \textit{[Europe, hasMajorCity, Munich]} is the city. The generated description is attached to the target entity and serves two purposes: it provides input for the description-context NED below, and it becomes the entity's description for all future decisions (e.g., during subsequent NED rounds).
\item \textbf{Description-Context NED:} For labels shared by a large number of entities (e.g., \textit{Faculty of Law} appears over 300 times, one per university), Triple-Context NED's candidate retrieval cannot surface all homonymous candidates, and the correct match may be absent from the candidate set. Details of an ablation are provided in Appendix~\ref{app:ablationonned}. Hence, this round is a focused pass that handles such high-density homonyms. The candidate set is restricted to entities sharing the target's exact label, and among these, the $k$ most similar candidates are retrieved by description embedding similarity. The LLM is then prompted with the target's generated description instead of the source triple as context (Fig.~\ref{fig:prompt-ned2}) and selects either a matching candidate or \textit{None of the above}.
\end{enumerate}

\paragraph{Guarded Parallelization} Sequential NED is computationally prohibitive at scale, but naive parallelization risks severe duplication. Consider, for instance, the entities \textit{Munich, Germany} in \textit{[BMW, foundingLocation, Munich, Germany]} and \textit{Munich} in \textit{[Europe, hasMajorCity, Munich]}. If neither currently exists in the KB and both are processed concurrently, they cannot be retrieved as candidates for one another; both are then classified as \textit{None of the above} and redundantly committed as new entities. We propose a guarded parallelization strategy that defers entities whose duplicates may not yet exist as candidates, while parallelizing the rest.
The strategy operates on the batching of pending NED requests, illustrated in Fig.~\ref{fig:architecture}. We retrieve all triples whose objects are pending NED and cluster them by object label embedding. From each cluster, we select the object label with the earliest elicitation timestamp (i.e., the entity most likely to be already established in the KB) and apply a frequency rule: if this label has already undergone NED more than a predefined threshold, all triples sharing this object label are added to the current batch, since the entity is likely already a candidate in the KB and can be safely processed in parallel. Otherwise, only the single earliest-elicited triple with this object label is included in the current batch; the remaining triples are deferred to subsequent batches, by which point the first triple's NED will have either matched the entity to an existing candidate or committed it as new, allowing the deferred triples to find it as a candidate.

\paragraph{Classes and Relations}
On-the-fly disambiguation is also applied to relations and classes (the objects of \textit{instanceOf} triples). Relations and classes still exhibit synonymy, e.g., \textit{isLocatedIn} and \textit{locatedIn}, or \textit{Film} and \textit{Movie}. But homonymy is rare and the candidate space is far smaller than for entities. The pipeline for these elements is therefore simplified to a single round of Triple-Context NED followed by description generation, with \textit{Unsure} omitted from the fallback options since the smaller candidate space makes deferring a decision unnecessary. Example prompts are shown in Figs.~\ref{fig:prompt-pd}--\ref{fig:prompt-cdg}.

\subsection{Caching}
To further reduce computational cost, we cache disambiguation decisions for elements that recur frequently, bypassing redundant NER and disambiguation calls. We apply two distinct caching policies depending on the element type. For entities, the cache activates if the (relation label, object label) pair from the source triple has occurred more than $\lambda$ times and all historical occurrences of the objects have linked to the same entity ID. The cache is checked at two points in the pipeline: before NER (allowing the object to skip both NER and NED) and, if NER classifies the object as a named entity, again before NED. The second check exists because concurrency may change the cache state between NER and NED.
For relations and classes, because relations and classes exhibit much less homonymy, a simple existence-based policy suffices: once a relation or class label is added to the KB, subsequent occurrences of the same label are cached and linked directly to the existing element.

\section{Experiments}

\subsection{Model and Hyperparameter Selection}

\paragraph{LLM Selection}
Our pipeline contains four LLM-driven tasks: knowledge elicitation, NER, description generation, and disambiguation. Stronger models are expected to improve performance on all four tasks, but at a trade-off with cost. We therefore compared three commercial models, GPT-5.1, GPT-5-mini, and GPT-5-nano, on task-specific samples. Based on this comparison, we use GPT-5.1 for elicitation and description generation, where output breadth and quality matter most, and GPT-5-mini for NER and NED, where it is competitive to the larger model at substantially lower cost. Full experimental details are in Appendix~\ref{app:model-selection}.

\paragraph{Hyperparameter Selection}
Similar trade-offs apply to the other parameters. For embedding-based candidate retrieval and NE clustering, we selected Qwen3-Embedding-4B. We retrieve the top-$5$ candidates for each NED decision. The thresholds for NED parallelization and caching were both set to 50 from development samples to preserve precision while still amortizing repeated high-frequency decisions. In all cases, we favored settings that keep the system scalable without materially degrading canonicalization quality. Details of experiments on hyperparameter selection are in Appendix~\ref{app:hyperparameter-selection}.

\paragraph{Seed Selection}
Mirroring \cite{gptkbv1}, all experiments start from the seed entity \entity{E0}{Vannevar Bush}. This choice is arbitrary and not conceptually important. General world knowledge forms a densely connected semantic graph, so recursive expansion from many reasonable seeds reaches the same large connected region~\cite{steyvers-tenenbaum-2005} quickly, e.g. via central entities such as New York, USA, or World War II one hop away. This is consistent with studies of Wikipedia, which find a giant connected component containing more than 93\% of nodes~\cite{ruprechter-santos-helic-2020-quality-links}. Seed irrelevance has also been empirically confirmed for recursive knowledge elicitation~\cite{giordano-razniewski-2026-foundations}. The seed is therefore a convenient starting point rather than a substantive modeling choice.

\begin{table}[t]
\centering
\small
\begin{tabular}{@{}lr@{}}
\toprule
\textbf{Entities} & 1,592,185   \\
\textbf{Triples} & 38,450,135 \\
\textbf{Relations} & 207,633 \\
\textbf{Classes} & 66,523 \\
\textbf{Triple objects (entities)} & 15.8M \\
\textbf{Triple objects (literals)} & 22.6M \\
\textbf{Avg.\ triples per entity} & 38.4 \\
\textbf{Avg.\ outlinks per entity} & 15.8 \\
\textbf{Entities novel to Wikidata}* & 36.8\% \\
\textbf{LLM API cost} & \$ 6,992 \\
\bottomrule
\multicolumn{2}{l}{\small\textbf{*} \textit{Validated on 1,000 samples.}}\\
\end{tabular}
\caption{Overall statistics of \ourkb.}
\label{tab:statistics}
\end{table}

\subsection{Large-scale Execution}
Using the above configuration, we run the full pipeline to construct \ourkb. As shown in Table~\ref{tab:statistics}, the resulting KB contains 38.4M triples and 1.6M disambiguated entities, together with 207.6K relations and 66.5K classes. Notably, based on a random sample of 1,000 canonical entities, 36.8\% are novel to Wikidata. We analyze this long-tail phenomenon  further in Section~\ref{longtail}. The full run took 70 days of active construction time. In total, it required 47.7M API calls and incurred a cost of \$6,992. Further analysis of construction scaling dynamics is available in Section~\ref{subsec:scaling} and Appendix~\ref{app:tokenusage}.

\section{Evaluation}

\subsection{Disambiguation Performance}
\label{subsec:ned-eval}
Disambiguation has a significant effect on the structure of the resulting KB. We manually evaluate NED quality along two directions of error: \emph{merge accuracy}, and \emph{split accuracy}. Results are shown in Table~\ref{tab:evaluation}. 

\paragraph{Merge Accuracy}
For \textit{homonym merges}, we sample 100 cases where an elicited triple object was merged into an existing entity with the same surface label and manually verify whether the two refer to the same real-world entity. 98 of 100 cases (98\%) are correct; the remaining 2\% are homonymy errors in which the pipeline failed to distinguish two same-label but distinct entities. For \textit{synonym merges}, we sample 100 triples whose object is an entity (excluding literal objects) and whose surface label differs from the canonical label of the entity it was mapped to, i.e., cases where an alias was resolved to an existing entity rather than added as new. We manually verify whether each alias maps to the correct real-world entity. Merge accuracy is 91 of 100 (91\%); the remaining 9\% are false merges in which an alias was incorrectly resolved to a different entity.

\paragraph{Split Accuracy}
For \textit{homonym splits}, we sample 100 entity pairs that share the same label but are assigned distinct IDs and manually verify whether each pair refers to two different real-world entities. All 100 pairs (100\%) are confirmed distinct, indicating no false splits in our sample.
For \textit{synonym splits}, exhaustively checking the entire KB for synonymous duplicates is infeasible, so we approximate through targeted sampling. For each of 100 random entities, we retrieve the top-20 nearest neighbors by label embedding similarity and manually check whether any candidate refers to the same real-world entity. For 95 of 100 samples (95\%), no synonymous duplicate is found. This estimate captures only duplicates whose labels are embedding-similar to the sampled entity; duplicates with very different surface forms may go undetected, so the true synonym split accuracy could be slightly lower.

\paragraph{Relations and Classes} Disambiguation quality for relations and classes is also evaluated, following the same setup as for entities. Relations achieve 89\% merge and 91\% split accuracy; classes achieve 87\% merge and 94\% split accuracy. The high split accuracy indicates that most fine-grained relations and classes are genuinely distinct from existing ones rather than un-merged duplicates. Inspection shows that many high-count relations arise from the LLM encoding n-ary or context-specific information as parameterized binary predicates: for example, \textit{distanceToDresden} (used in triples such as (\textit{Chemnitz}, \textit{distanceToDresden}, \textit{75km})) encodes a ternary distance relation with one argument embedded in the relation name, and \textit{adjacentTimeZoneWest} similarly parameterizes a directional relation. These are structural effects of representing n-ary knowledge in a binary triple form, not disambiguation failures. Some residual under-canonicalization nonetheless remains, where semantically equivalent relations or classes could ideally be merged further.

\paragraph{Summary} The pipeline reliably handles homonymy in both directions and synonymy with high precision overall, with the largest error mode being false merges in synonymy resolution (9\%). Even extreme homonymy cases are correctly resolved: 309 distinct entities labeled \textit{Faculty of Law} and 113 labeled \textit{Terminal 1} are maintained as separate entities. Synonymy resolution consolidates large alias sets, with \entity{E14}{United States of America} reaching 128 aliases as the most extreme case. Further details on the evaluation, including annotator reasoning, are in Appendix~\ref{app:disambiguation-evaluation}.


\subsection{Comparison to Baseline}

We compare \ourkb to a baseline that uses surface forms for entity identification. At 2,316,275 entities, the baseline contains a substantial number of duplicates (vs. 1,592,185 in \ourkb), and at the same time, conflates 131,990 entities that share one of 41,275 surface labels with others.
For example, in the baseline, there are a total of 365 entities that are \predicate{P10}{memberOf} the \entity{E2721}{Eurozone}, while in \ourkb, there are only 30.\footnote{In reality, there are 21 countries, the overcount being explained by some modeling complexity of distinguishing countries and governments in \ourkb.}


In addition, we performed a focused comparison of the two approaches: In the style of \cite{giordano-razniewski-2026-foundations}, we executed both the baseline and \ourmethod on a specific domain, Ancient Babylon, both up to a size of 2,700 entities.
In Babylon-DisamKB, 883 entities are associated with 4,670 surface labels, indicating substantial synonymy. Of these 883 entities, 447 also appear in Babylon-BaselineKB. The remainder are not yet reached by the BFS, highlighting the inefficiency of non-disambiguated knowledge extraction. 
Even homonymy occurs already at this scale: The labels \textit{Paris} (city and Trojan prince), \textit{Philadelphia} (Greek and US city) and \textit{Zechariah} (different persons) are wrongly conflated by the baseline. 

\begin{table}
\centering
\small
\begin{tabular}{@{}lcc@{}}
\toprule
\textbf{Judge} & \textbf{Human} & \textbf{LLM} \\
\midrule
\multicolumn{3}{@{}l}{\textbf{Disambiguation accuracy}} \\
\quad \textbf{Entity} &  &   \\
\quad Merge correct & 94.5\% & --- \\
\quad Split correct & 97.5\% & --- \\
\quad \textbf{Relation} &  &   \\
\quad Merge correct & 89\% & --- \\
\quad Split correct & 91\% & --- \\
\quad \textbf{Class} &  &  \\
\quad Merge correct & 87\% & --- \\
\quad Split correct & 94\% & --- \\
\addlinespace
\multicolumn{3}{@{}l}{\textbf{Triple factuality}} \\
\quad True & 94.5\% & 92.8\% \\
\quad Plausible & 2.5\% & 5.5\% \\
\quad Implausible & 1.0\% & 1.2\% \\
\quad False  & 2.0\% & 0.5\% \\
\addlinespace
\multicolumn{3}{@{}l}{\textbf{Entity factuality}} \\
\quad Verifiable & 96.0\%/90.5\%* & 92.3\% \\
\quad Plausible & 2.0\%/4.9\%* & 6.7\%  \\
\quad Unverifiable & 2.0\%/4.6\%* & 1.0\%  \\
\bottomrule
\multicolumn{2}{l}{\small\textbf{*} \textit{On the subset of entities novel to Wikidata}}\\
\end{tabular}
\caption{Manual and automatic evaluation of the constructed KB. Human evaluation uses $n=400$ for Disambiguation accuracy and $n=200$ for triple and entity factuality; LLM evaluation uses $n=1{,}000$.}
\label{tab:evaluation}
\end{table}

\subsection{Overall KB Quality}
Exhaustive evaluation of a 38M-triple KB is infeasible, so we evaluate randomly sampled subsets. Following \cite{yago} and recent metrics like FactScore \cite{min-etal-2023-factscore} and VeriScore \cite{song-etal-2024-veriscore}, we use web-retrieved snippets as the evidence source for verification. We conduct two parallel evaluations using the same labeling scheme: a smaller sample of 200 items annotated by human, and a larger sample of 1,000 items automatically evaluated by an agentic pipeline with web search. We find a precision of $>$90\% for all of NED, triples, and entities. The full breakdown for both triples and entities is in Table~\ref{tab:evaluation}, and further details on the evaluation, including agreement, are in Appendix~\ref{app:overall-evaluation}.
\paragraph{Triple factuality.} Each triple is labeled as one of: \textit{true} (directly supported by web sources), \textit{plausible} (consistent with web sources but not directly attested), \textit{implausible} (inconsistent), or \textit{false} (contradicted). By the human judge, 94.5\% of triples are labeled true and 2.0\% false; by the LLM judge, 92.8\% are true and 0.5\% false, with the remainder distributed across plausible and implausible. 
Further details are in Appendix~\ref{app:overall-evaluation}.

\paragraph{Entity factuality.} Each entity is labeled as one of: \textit{verifiable} (corresponds to a real-world entity confirmed by web sources), \textit{plausible} (likely real but not confirmed), or \textit{unverifiable} (no supporting evidence found). Because entity verification is difficult without context, we provide the entity's description to evaluators as supporting information. On the manual sample, 96\% of entities are verifiable and 2\% unverifiable; on the LLM sample, 92.3\% are verifiable and 1\% unverifiable. Furthermore, we also verified entity factuality of all entities in subset that novel to Wikidata. We find that 90.5\% are verifiable, 4.9\% plausible, 4.6\% unverifiable. This is modestly lower than the KB-wide rate, consistent with mildly elevated hallucination risk in the novel subset. Nonetheless, the large majority of novel entities are verifiable, indicating that most out-of-Wikidata entities correspond to real-world knowledge.

\section{Discussion}

\subsection{Potential and Enabled Analyses}

\ourkb provides significant novelty to KBC, for the first time providing a large, disambiguated entity repository outside the Wikimedia domain. We analyzed the fraction of entities in \ourkb that cannot be found in Wikidata, finding a staggering 36.8\% that are novel.

We provide an online browsing interface for \ourkb at \url{https://gptkb.org/}, including unique web links to entities (e.g., {\small\url{https://gptkb.org/entity/E319/}} for \entity{E319}{Martin Luther King}). This enables both a per-entity visualization of disambiguation decisions, elicited triples, and aliases, as well as aggregate query capabilities via a SPARQL endpoint. Aggregate queries, in turn, enable large-scale statistical analysis, which can illuminate aspects such as topical foci, gender and nationality biases, logical consistency, timeliness, and similar \cite{ghosh2025mining}. For example, in the web interface we provide SPARQL queries that highlight clear gender (70k male, 34k female), or region biases (86k USA vs.\ 1k China).

\subsection{Long-tail Coverage Beyond Wikidata}
\label{longtail}

Two observations characterize the long-tail nature of \ourkb. 
First, most entities are long-tail: 1,330,135 entities (83.6\% of canonical entities) occur fewer than five times, and 885,218 (55.6\%) occur only once. Second, to estimate overlap with Wikidata, we randomly sampled 1,000 canonical entities from \ourkb and validated their existence in Wikidata. 36.8\% are not present, indicating substantial coverage beyond curated KBs. Among these Wikidata-novel entities, 89\% occur fewer than five times in \ourkb and 62.5\% occur only once. These rates exceed the corresponding overall rates, so divergence from Wikidata is concentrated in the long-tail region of \ourkb. These observations point to a key advantage of LLM-based KB construction: LLM-derived KBs can readily extend coverage into the sparse long-tail region where curated effort is least developed, complementing curated resources rather than merely accelerating traditional KB construction.

\begin{figure}[t]
    \centering
    \includegraphics[width=1\linewidth]{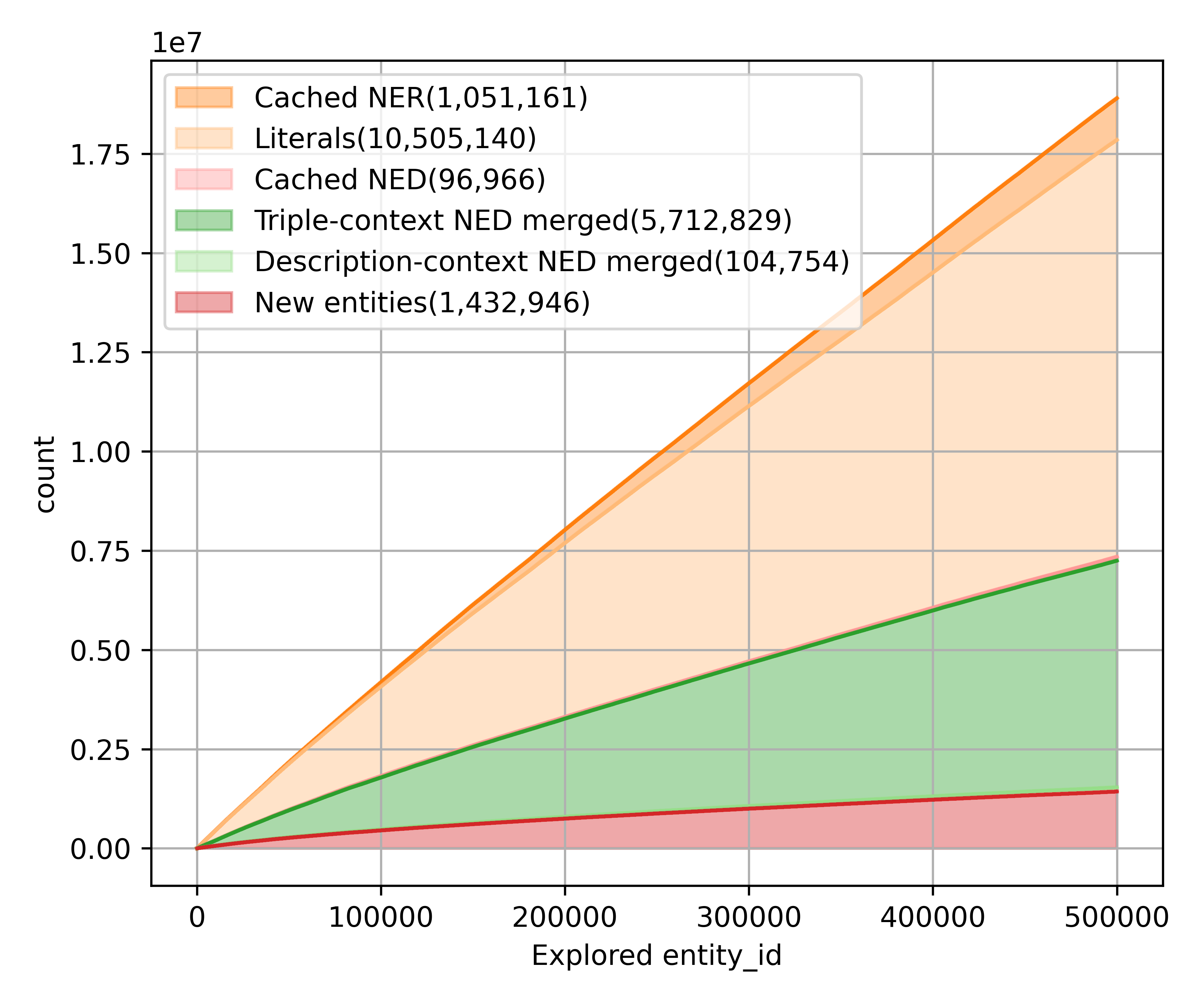}
    \caption{Scaling behaviour over construction.}
    \label{fig:scaling}
\end{figure}

\subsection{Scaling Behaviour}
\label{subsec:scaling}
To understand the dynamics of our on-the-fly construction pipeline, we analyze how the composition of triple objects evolves as the KB grows. Figure~\ref{fig:scaling} shows the cumulative breakdown of object resolutions as exploration expands to 500K entities. As the KB grows, the curves show sublinear scaling, but relative composition of object resolutions remains stable. Excluding literal objects (10.5M), the pipeline merges entity objects into existing entities far more often than it instantiates them as new: 5.7M merges via Triple-Context NED and 105K via Description-Context NED, against 1.4M new entities. This indicates that the pipeline builds a densely connected graph rather than an expanding set of disconnected new entities. Caching avoids API calls for 1.2M triple objects: 1.1M before NER (each skipping both NER and NED) and further 100K before NED, with savings growing steadily as the KB expands. Additional analysis of token consumption and cost dynamics are available in Appendix~\ref{app:tokenusage}.

\section{Conclusion}
In this paper, we introduced a novel methodology that enables context-enhanced, on-the-fly disambiguation during the recursive materialization of LLM knowledge. By effectively resolving the pervasive challenges of synonymy and homonymy, our approach significantly improves the precision and scalability of LLM-based KB construction. Along with this work we release GPTKB 2.0, a 38M triple epistemic twin of GPT 5.1.

\section{Limitations}

Our work has several limitations. First, \ourkb currently does not attach references or provenance information to individual triples. While triples can be manually inspected or post-hoc verified, the KB itself does not provide evidence for why a fact should be trusted. Adding references retrospectively is possible, but remains challenging at scale.

Second, despite substantial computational effort and monetary cost, \ourkb remains much smaller than Wikidata. This comparison should be interpreted cautiously, since Wikidata contains many highly templatic records, including a large share of bibliographic statements, and the efforts behind it are in the multi-million monetary equivalent \cite{paulheim2018much}. Still, many long-tail entities and facts are likely missing.

Third, \ourkb is time-static. Its facts reflect the knowledge of the underlying models at training time, including its cutoff, outdated beliefs, and errors. Updating the KB requires re-running the pipeline with a newer model or adding dedicated mechanisms for detecting and refreshing stale facts \cite{10.1145/3774904.3792695}.

Overall, \ourkb should be viewed not as a gold-standard KB, but as a materialized approximation \cite{razniewski_2026_19910437} of LLM's factual beliefs.


\section{Ethical considerations}
Since the statements in our knowledge base (KB) are entirely elicited from the parametric knowledge of LLMs, they may inadvertently inherit undesirable information, such as social biases or toxicity, that was not fully mitigated during the models' training.

\bibliography{references}

\appendix
\section*{Appendix}

\section{Experiment Setups}
\subsection{Model Selection Details}
\label{app:model-selection}

\begin{table*}
\centering
\begin{tabular}{@{}ll|rrr@{}}
\toprule
\textbf{Task} & \textbf{Dataset} & \textbf{GPT-5.1} & \textbf{5-mini} & \textbf{5-nano} \\ \midrule
\multirow{1}{*}{\textbf{\begin{tabular}[c]{@{}l@{}}Elicitation (Med.\ \#triples)\end{tabular}}} &  & 47 & 33 & 8 \\
True & & 0.835 & 0.705 & \textbf{0.865} \\ 
 \midrule
\multirow{3}{*}{\textbf{\begin{tabular}[c]{@{}l@{}}NER (accuracy)\end{tabular}}} & Full set & 0.93 & \textbf{0.96} & 0.79 \\
 & \textit{- entities} & 0.87 & 0.93 & 0.98 \\
 & \textit{- literals} & 1.0 & 1.0 & 0.57 \\ \midrule
\multirow{3}{*}{\textbf{\begin{tabular}[c]{@{}l@{}}NED (accuracy)\end{tabular}}} & Full set & \textbf{0.99} & \textbf{0.99} & 0.86 \\
 & \textit{- split} & 0.98 & 0.98 & 0.80 \\
 & \textit{- merge} & 1.0 & 1.0 & 0.92 \\ \bottomrule
\end{tabular}
\caption{Model comparison on the three main tasks.}
\label{tab:model-comparison}
\end{table*}

We compared GPT-5.1, GPT-5-mini, and GPT-5-nano on three task-specific evaluations: elicitation, NER, and NED. Results are in Table~\ref{tab:model-comparison}.

\paragraph{Elicitation}
We elicited triples for 100 randomly sampled entities using the prompt in Fig.~\ref{fig:prompt-elicitation}. We compared the number of generated triples quantitatively and manually assessed 200 sampled triples (without subject description as context) for quality.

\paragraph{NER}
We constructed 100 NER requests using the prompt in Fig.~\ref{fig:prompt-ner} and evaluated outputs against human-annotated ground truth.

\paragraph{NED}
We constructed 100 NED requests using the prompt in Fig.~\ref{fig:prompt-ned1}, with 50 \textit{Split} cases (subset of samples with ground truth that to add as new entity) and 50 \textit{Merge} cases (subset of samples with ground truth that to merge to existing entity), and compared model predictions against human annotation.

GPT-5.1 generated a median of 47 triples per entity, compared to 33 for GPT-5-mini and 8 for GPT-5-nano. Triple precision was 0.835 for GPT-5.1 and 0.865 for GPT-5-nano, but the much broader factual coverage of GPT-5.1 made it the preferable choice for elicitation. For NER, GPT-5-mini achieved 0.96 accuracy, compared to 0.93 for GPT-5.1 and 0.79 for GPT-5-nano. For NED, GPT-5-mini and GPT-5.1 both achieved 0.99 overall accuracy, substantially above GPT-5-nano at 0.86. Because GPT-5-mini matches GPT-5.1 on NER and NED while being much cheaper, we use GPT-5-mini for both tasks and GPT-5.1 for elicitation and description generation.

\subsection{Hyperparameters Selection}
\label{app:hyperparameter-selection}
\paragraph{Embedding Model}
We compared three embedding models from the Qwen-3-Embedding family (the 0.6B, 4B, and 8B variants) on manually curated test set of 20 difficult disambiguation cases. Each case contained five entities, with known semantic relationships (some intended to be close, others distant), e.g., [\textit{Treaty of Paris (1763)}, \textit{Treaty of Hubertusburg (1763)}, \textit{Paris}, \textit{Treaty of Paris}, \textit{Treaty of Baden (1714)}]. For each model, we computed pairwise similarity scores within each set and manually assessed whether the resulting rankings reflected the intended relationships. We found that Qwen-3-Embedding-4B and Qwen-3-Embedding-8B produced highly similar scores and reliably distinguished close from distant entity pairs, while Qwen-3-Embedding-0.6B did not. Given the comparable quality of the two larger models, we adopted Qwen-3-Embedding-4B for all embedding tasks in our pipeline (label and description embeddings for candidate retrieval, and clustering for parallelization) as a computational efficiency choice.

\paragraph{Disambiguation Candidate Number $k$}
We retrieve the top-5 candidates for each NED decision, as preliminary inspection showed that the correct target is concentrated heavily to the first candidate with 88\%, to the second with 2.5\% and 0 to the 5th candidate among subset of 50 samples in NED performance evaluation shown in Appendix~\ref{app:model-selection}. This indicates that to select 5 most possible candidates is sensible to save cost while ensuring that the correct candidate is included.

\paragraph{Caching and Guarded Parallelization Thresholds}
We set the frequency threshold to 50 for both caching and guarded parallelization, based on development samples. To verify that this threshold preserves precision, we conducted two manual evaluations on a KB constructed to a scale of 100K entities.
For caching, we sampled 100 triples in which the object was directly linked to an existing entity through caching (skipping NED), and manually verified each merge decision. All 100 merges (100\%) are correct, indicating that a threshold of 50 is sufficient to ensure that cached pairs unambiguously identify a single entity.
For guarded parallelization, we sampled 100 entity pairs that share the same surface label and were processed in the same NED batch, and manually verified that the correct existing candidate appeared in the candidate set for both members of each pair. All 100 pairs (100\%) satisfy this condition, confirming that the threshold prevents the concurrency failure mode in which two co-occurring duplicates each fail to retrieve the other.
A lower threshold might preserve precision while yielding additional efficiency gains; we did not sweep this hyperparameter, since the goal of these mechanisms is to enable scalable construction while maintaining canonicalization quality, both of which are already achieved at 50.

\section{Ablation Study on the NED Module}
\label{app:ablationonned}
To evaluate the contribution of the description-context NED step to disambiguation performance, we constructed an ablated KB using a pipeline variant that uses only the first phase, triple-context NED. Apart from this single change, all components, including model selection and hyperparameters, remain unchanged. The ablated KB contains 2.3M triples.
We sampled 100 \textit{split} cases (where the ground-truth decision is to add the entity as new) and 100 \textit{merge} cases (where the ground-truth decision is to merge with an existing entity), and manually evaluated each.
On the split subset, the ablated pipeline achieves 95\% precision: 5 of 100 entities that should have been added as new were instead incorrectly merged to existing entities. These are false merges that the second round could have caught using description-based context.
On the merge subset, the ablated pipeline merged each of the 100 samples into an existing entity, but inspection of the merged-into entities revealed that in 22\% of cases, additional duplicate entities for the same real-world referent already existed in the KB, duplicates that earlier triple-context decisions failed to consolidate. In contrast, the full pipeline has less than 5\% false splits (cf. Section~\ref{subsec:ned-eval}). This indicates that without the focused homonymy check in round 2, duplicates accumulate over the course of construction.

\section{Cost Dynamics and Token Consumption} 
\label{app:tokenusage}

\begin{figure}
    \centering
    \includegraphics[width=1\linewidth]{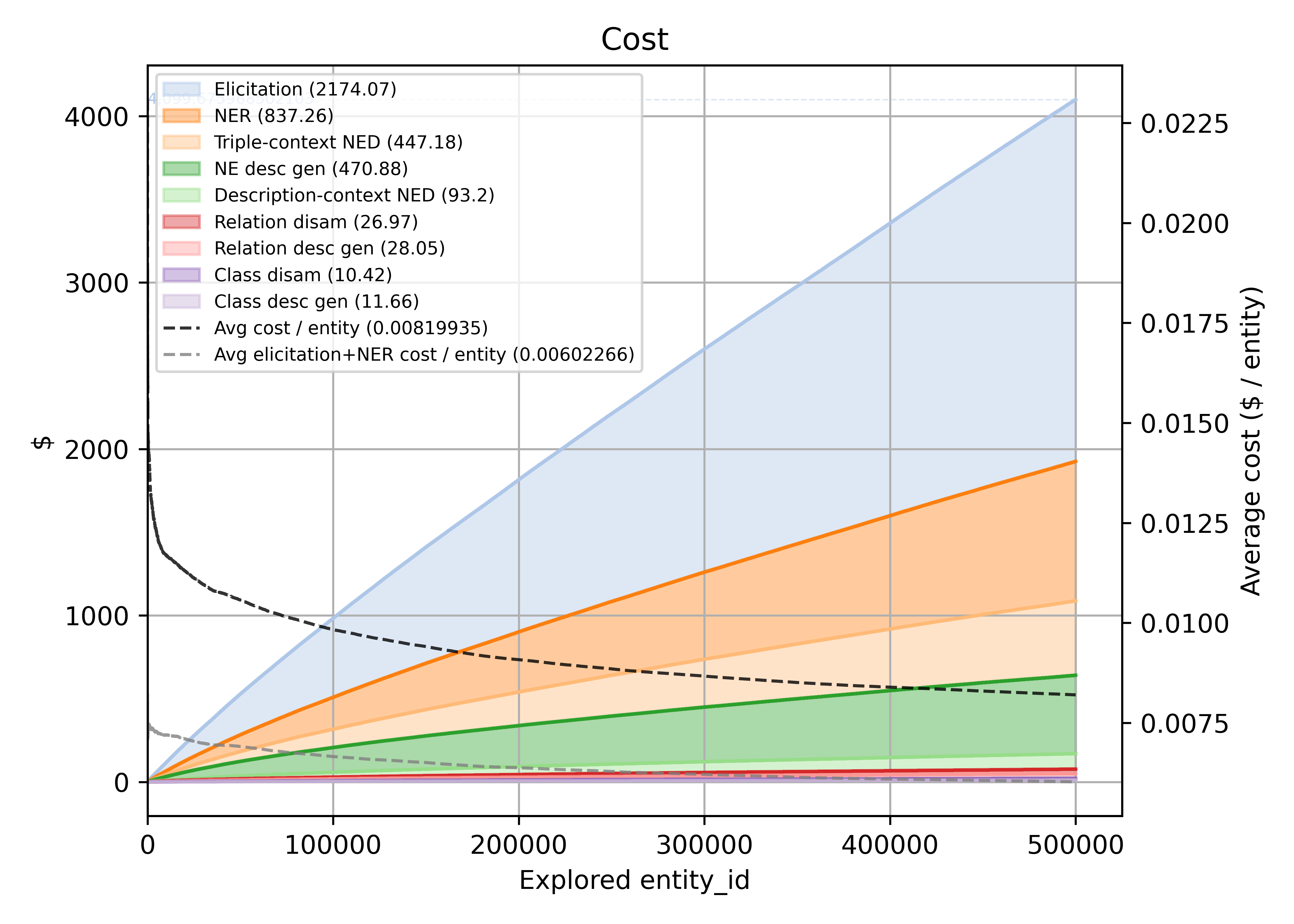}
    \caption{Cost dynamics over the first 500K explored entities.}
    \label{fig:cost-dynamic}
\end{figure}

Figure~\ref{fig:cost-dynamic} illustrates the cost dynamic of all phases during exploration of 500K entities. While cumulative costs grow predictably, they are overwhelmingly dominated by the Elicitation (\$2174) and NER (\$837) phases. As depicted by the black dashed line in the plot, the \textit{overall average cost per entity} exhibits a sharp initial decline and steadily decreases as the KB scales, dropping from \$0.015 toward \$0.008. This decreasing marginal cost demonstrates the profound benefit of our context-enhanced disambiguation paradigm and caching strategy: a larger, denser KB yields higher cache hit rates and more frequent entity merges, effectively bypassing expensive downstream operations like the two rounds of NED and new description generation. Further more, Tabel~\ref{tab:costv2} shows the cost and token distribution of LLM calls on model-based tasks across pipeline phases. The dominant expenses come from elicitation and entity-centric processing. Elicitation consumes relatively few calls but is costly because it uses GPT-5.1 and produces long outputs; in our run, it accounts for 59.5\% of total cost. Entity-related operations, especially NER and NED, dominate token consumption and together account for 38.8\% of total cost. By contrast, class and relation canonicalization contribute only a small fraction of overall spending. 

Figure~\ref{fig:cost-dynamic} illustrates cost dynamics across all phases over the first 500K explored entities. Cumulative costs grow approximately linearly with the number of explored entities and are dominated by the Elicitation (\$2,174) and NER (\$837) phases. The black dashed line shows the \textit{overall average cost per entity}, which declines sharply for the first ~50K entities (amortization of initial high-cost entities) and continues to decline gradually thereafter (efficiency gains from caching and consolidation as the KB grows), stabilizing near \$0.008 per new entity by 500K entities.
Table~\ref{tab:costv2} reports the cost and token distribution across pipeline phases, aggregated over the construction (1M entities explored). Elicitation accounts for the largest share of total cost (59.5\%), because each call uses the high-capacity model and produces long outputs. NER and entity disambiguation steps together account for 38.8\%, with NER dominating call volume (33.7M calls). Class and relation canonicalization together contribute under 2\% of total cost, which is consistent with their simpler single-round design.

\begin{table}
\resizebox{\columnwidth}{!}{
\begin{tabular}{@{}lrrrr@{}}
\toprule
 & \textbf{API cost} & \textbf{Calls} & \textbf{Input tokens} & \textbf{Output tokens}\\ \midrule
\textbf{Elicitation} & \textbf{59.5\%} & \textbf{1M} & \textbf{265.7M} & \textbf{798.8M}\\
\textbf{Entities} & \textbf{38.8\%} & \textbf{45.3M} & \textbf{7.8B} & \textbf{1.3B} \\
\textit{\quad NER} & \textit{22.6\%} & \textit{33.7M} & \textit{4.2B} & \textit{1B}  \\
\textit{\quad Triple-context NED} & \textit{7.2\%} & \textit{8.2M} & \textit{2.8B} & \textit{156.3M} \\
\textit{\quad Description Gen} & \textit{7.5\%} & \textit{1.7M} & \textit{228.6M} & \textit{76.3M} \\
\textit{\quad Description-context NED} & \textit{1.5\%} & \textit{1.7M} & \textit{569.7M} & \textit{32.5M} \\
\textbf{Classes} & \textbf{0.5\%} & \textbf{366.4K} & \textbf{94.6M} & \textbf{8.6M} \\
\textit{\quad Class Disam} & \textit{0.2\%} & \textit{300K} & \textit{91.2M} & \textit{5.4M}\\
\textit{\quad Description Gen} & \textit{0.3\%} & \textit{66.4K} & \textit{3.4M} & \textit{3.2M} \\
\textbf{Relations} & \textbf{1.2\%} & \textbf{1M} & \textbf{219.3M} & \textbf{22.3M} \\ 
\textit{\quad Relation Disam} & \textit{0.6\%} & \textit{823.5K} & \textit{205.4M} & \textit{15.7M} \\
\textit{\quad Description Gen} & \textit{0.6\%} & \textit{207.7K} & \textit{13.9M} & \textit{6.6M} \\ \bottomrule
\end{tabular}
}
\caption{Cost and token distribution of LLM calls for KB construction.}
\label{tab:costv2}
\end{table}

\section{Evaluation Details}

\subsection{Disambiguation}
\label{app:disambiguation-evaluation}
Manual annotation for the NED evaluation (n=400 in total) covers four subsets corresponding to the two phenomena (homonymy and synonymy) and the two directions of error (merge and split), with 100 samples each. Examples from each subset are provided in Tables~\ref{tab:human-eval-homonymy-merge}--\ref{tab:human-eval-synonymy-split}.
\paragraph{Homonymy merges (n=100).} The annotator manually verified whether the merged triple object and entity-to-merge refer to the same real-world entity. Correct merges are labeled \textit{True merge}; incorrect merges (where two distinct entities sharing a label were wrongly combined) are labeled \textit{False merge}.
\paragraph{Synonymy merges (n=100).} Similar to homonymy merges, but here the triple object's surface label differs from the label of the entity-to-merge (i.e., the object was treated as an alias). The annotator verified whether the alias resolution was correct. Correct mappings are labeled \textit{True merge}; incorrect mappings are labeled \textit{False merge}.
\paragraph{Homonymy splits (n=100).} The annotator verified whether entity A and entity B in each sample pair refer to distinct real-world entities. Correct splits are labeled \textit{True split}; incorrect splits (where two same-label mentions of the same entity were wrongly kept apart) are labeled \textit{False split}.
\paragraph{Synonymy splits (n=100).} The annotator checked whether any entity in the candidate list refers to the same real-world entity as the sample. Samples with no duplicate found are labeled \textit{True split}; samples with a duplicate found are labeled \textit{False split}.

\subsection{Overall Quality}
\label{app:overall-evaluation}
For triples (n=200), the human evaluation was performed by an author of this paper, with a subsample (n=100) annotated by a second author. On that subsample, the two annotators agreed on 90/100 rows. Overall true rate based on the annotators was  94.5\% / 90\%, overall false rate was 2\% / 1\%, with the remainder plausible/implausible. Human evaluation examples of triples and entities are provided in Tables.~\ref{tab:human-eval-triple-precision}--\ref{tab:human-eval-entity-factuality}.

For the automatic evaluation, we compared two settings: In the default setting, the LLM judge was provided the triple and the subject description, and was tasked, in an agentic manner, to first perform web search for evidence (using the Brave Search API), then in a second step, judge triple truth based on the web evidence (column ``LLM'' in Table~\ref{tab:evaluation}).
Entity descriptions help with understanding the meaning of a triple, however, introduce also a risk of biasing the system by content created by an LLM itself. 

We also compared three LLM judges with each other, Llama-4-Scout with fixed template-based queries and Opus 4.7 and Gemini 2.5 Pro in the agentic setting. We find a high agreement in true rate on the bare task (79.9\%/76.2\%/83.9\%), and a solid agreement also for the other labels for the two agentic models, but substantial variation from them to Llama-4-Scout: Plausible: 0\% versus 19.2\%/11.7\%, false 19.9\% versus 1.7\%/1.4\%). This indicates that the ability to formulate custom queries is an important part of the automated judge protocol. 

\section{Prompts}
Prompts we used in LLM-based phases are provided in Figures.~\ref{fig:prompt-elicitation}--\ref{fig:prompt-cdg}.


\begin{table*}
\centering
\small
\begin{tabular}{@{}p{2cm} p{2.5cm} p{1.5cm} p{1.5cm} p{4cm} p{4cm}@{}}
\toprule
\textbf{Subject} & \textbf{Relation} & \textbf{Object} &\textbf{Entity-to-merge} & \textbf{Description of entity-to-merge} & \textbf{Evaluation}\\ 
\midrule
Aldebaran planetary system & locatedInConstellation & Taurus & Taurus & Taurus is a prominent zodiac constellation in the northern sky, symbolized as a bull and known for containing the bright star Aldebaran and the Pleiades star cluster. & \textbf{True merge} (Triple object entity and entity-to-merge refer to identical entity)\\
\addlinespace
Guidobaldo II della Rovere & influencedBy & Eleonora Gonzaga & Eleonora Gonzaga	& Eleonora Gonzaga was an Italian noblewoman of the Gonzaga family who became Holy Roman Empress as the wife of Emperor Ferdinand II. & \textbf{False merge} (Triple object \textit{Eleonora Gonzaga} refers to Eleonora Gonzaga (1493–1550), who married Francesco Maria I della Rovere and was the mother of Guidobaldo II della Rovere, Duke of Urbino. Entity-to-mege \textit{Eleonora Gonzaga}, who married Holy Roman Emperor Ferdinand II in 1622, making her Holy Roman Empress. She lived a century later and belonged to a different generation of the Gonzaga family.\\
\bottomrule
\end{tabular}
\caption{Examples of human evaluation on homonymy merges.}
\label{tab:human-eval-homonymy-merge}
\end{table*}

\begin{table*}
\centering
\small
\begin{tabular}{@{}p{2.5cm} p{1.5cm} p{1.5cm} p{1.5cm} p{4cm} p{4cm}@{}}
\toprule
\textbf{Subject} & \textbf{Relation} & \textbf{Object} &\textbf{Entity-to-merge} & \textbf{Description of entity-to-merge} & \textbf{Evaluation}\\ 
\midrule
Ashmont–Braintree branch of the Red Line & operator & Massachusetts Bay Transportation Authority & MBTA	& The MBTA (Massachusetts Bay Transportation Authority) is the public transit agency that operates subway, bus, commuter rail, and ferry services in the Greater Boston area. & \textbf{True merge} (Triple object entity \textit{Massachusetts Bay Transportation Authority} and entity-to-merge \textit{MBTA} refer to identical entity)\\
\addlinespace
MATE Settings Daemon & uses & GSettings & dconf	& dconf is a low-level configuration system and settings storage backend commonly used by GNOME and other Linux desktop applications to manage user preferences. & \textbf{False merge} (Triple object entity \textit{GSettings} is the API that applications (like MATE Settings Daemon) program against, but entity-to-merge \textit{dconf} is typically the backend that GSettings uses to persist the data on Linux systems.)\\
\bottomrule
\end{tabular}
\caption{Examples of human evaluation on synonymy merges.}
\label{tab:human-eval-synonymy-merge}
\end{table*}

\begin{table*}
\centering
\small
\begin{tabular}{@{}p{2cm} p{4cm} p{2cm} p{4cm} p{4cm}@{}}
\toprule
\textbf{Entity A} & \textbf{Entity A description} & \textbf{Entity B} & \textbf{Entity B description} & \textbf{Evaluation}\\ 
\midrule
Brother Jack & "Brother Jack" is a soul-jazz album by organist Jack McDuff that helped establish his reputation as a leading Hammond B-3 player in the early 1960s. & Brother Jack & Brother Jack is a prominent leader of the Brotherhood in Ralph Ellison’s novel "Invisible Man," symbolizing manipulative political idealism and racial betrayal. & \textbf{True split} (Entity A \textit{Brother Jack} and entity B \textit{Brother Jack} refer to different entities.)\\
\bottomrule
\end{tabular}
\caption{Examples of human evaluation on homonymy splits.}
\label{tab:human-eval-homonymy-split}
\end{table*}

\begin{table*}
\centering
\small
\begin{tabular}{@{}p{2.5cm} p{5cm} p{6cm} p{2cm}@{}}
\toprule
\textbf{Entity} & \textbf{Entity description} & \textbf{Candidates} & \textbf{Evaluation}\\ 
\midrule
Beaufort Shelf & The Beaufort Shelf is a broad, shallow continental shelf region in the Arctic Ocean off the northern coast of Alaska and Canada, forming part of the transition between the nearshore Beaufort Sea and the deep Canada Basin. & 
\textbf{1}: Scotian Shelf: The Scotian Shelf is a broad, shallow continental shelf off Atlantic Canada ...
\textbf{2}: Newfoundland and Labrador shelf: The Newfoundland and Labrador shelf is a broad, shallow marine region off eastern Canada ...
\textbf{3}:...
\textbf{18}: West Caroline Basin: The West Caroline Basin is an oceanic basin in the western Pacific Ocean...
\textbf{19}: Burin-Grand Bank: Burin-Grand Bank is a provincial electoral district in Newfoundland and Labrador, Canada...
\textbf{20}: Faxaflói Bay: Faxaflói Bay is a large bay in southwest Iceland... & \textbf{True split} (no duplicate entity)\\
\addlinespace
Etiopía / Plaza de la Transparencia	& Etiopía / Plaza de la Transparencia is a major public square and transport hub in Mexico City that serves as a key station and terminus on the Metrobús Line 3. &
\textbf{1}: Plaza de la Transparencia: Plaza de la Transparencia is a public square in Mexico City associated with themes of openness...
\underline{\textbf{2}}: Plaza Etiopía: Plaza Etiopía is a well-known public square and transport hub in Mexico City...
\textbf{3}:...
\textbf{18}: Plaza del Congreso: Plaza del Congreso is a major public square in Buenos Aires...
\textbf{19}: Plaza de Panama: Plaza de Panama is a central open square in San Diego’s Balboa Park...
\textbf{20}: Plaza de la Ciudadanía: Plaza de la Ciudadanía is a prominent public square in central Santiago, Chile...
& \textbf{False split} (2. refers to same entity)\\
\bottomrule
\end{tabular}
\caption{Examples of human evaluation on synonymy splits.}
\label{tab:human-eval-synonymy-split}
\end{table*}

\begin{table*}
\centering
\small
\begin{tabular}{@{}p{2.5cm} p{1.8cm} p{2.5cm} p{7cm} p{2cm}@{}}
\toprule
\textbf{Subject} & \textbf{Relation} & \textbf{Object} & \textbf{Subject description} & \textbf{Evaluation}\\ 
\midrule
The Theory of Wages & mainSubject & unemployment & The Theory of Wages is a foundational economic work by John R. Hicks that analyzes how wages are determined within competitive labor markets and broader economic systems. & False \\
\addlinespace
Calusa Beach & locatedIn & Monroe County, Florida & Calusa Beach is a small, family-friendly sandy beach with calm, shallow waters located within Bahia Honda State Park in the Florida Keys. & True \\
\addlinespace
Victoria Street railway station & distanceFrom & Maitland railway station & Victoria Street is a railway station on the line between Newcastle and Maitland in New South Wales, Australia. & Implausible \\
\bottomrule
\end{tabular}
\caption{Examples of human evaluation on triple precision.}
\label{tab:human-eval-triple-precision}
\end{table*}

\begin{table*}
\centering
\small
\begin{tabular}{@{}p{3cm} p{10.5cm} p{3cm}@{}}
\toprule
\textbf{Entity} & \textbf{Entity description} & \textbf{Validation}\\ 
\midrule
Torres de Segre & Torres de Segre is a municipality in the comarca of Segrià in the province of Lleida, Catalonia, Spain. & Verifiable \\
\addlinespace
John Martin & John Martin was an American statesman who represented South Carolina in the Continental Congress during the Revolutionary era. & Plausible \\
\addlinespace
God and Fatherland & God and Fatherland is the official English-language motto of the Colombian Air Force, reflecting its emphasis on religious faith and national loyalty. & Unverifiable \\
\bottomrule
\end{tabular}
\caption{Examples of human validation on entity factuality.}
\label{tab:human-eval-entity-factuality}
\end{table*}

\begin{figure*}
    \centering
    \noindent
    \fbox{\small 
    \begin{minipage}{\textwidth}
        \textbf{developer} \\
        You are a knowledge base construction expert. Given a subject entity and a description of it, return factual statements that you know for the subject as a JSON list of dictionaries(triples), where keys must be "subject", "predicate" and "object". The number of facts may be very high, between 25 to 50 or more, for very popular subjects. For less popular subjects, the number of facts can be very low, like 5 or 10.\\
        \# Requirements\\
        - If you don't know the subject at all, return an empty list.\\
        - If the subject is not a named entity, return an empty list.\\
        - Include at least one triple where predicate is "instanceOf".\\
        - Do not get too wordy.\\
        - Separate several objects into multiple triples with one object.\\
        \ \\
        \textbf{user} \\
        Subject: [Vannevar Bush] \\
        Description of subject: [American electrical engineer and science administrator (1890~1974)]
    \end{minipage}
    }
    \caption{Prompt for knowledge elicitation.}
    \label{fig:prompt-elicitation}
\end{figure*}

\begin{figure*}
    \centering
    \noindent
    \fbox{\small 
    \begin{minipage}{\textwidth}
        \textbf{developer} \\
        Given a phrase, classify it is english named entity (e.g., persons, organizations, works of art) in Latin script, or not (e.g., literals, dates, URLs, verbose phrases). For disambiguation, the statement where the phrase occurs as object is also given. Please return a JSON object with 'phrase' (string, the phrase being analyzed) and 'is\_ne' (boolean, indicating whether the phrase is a Named Entity).\\
        \ \\
        \textbf{user} \\
        Phrase: [Buenos Aires] | Statement: [Argentina, capital, Buenos Aires]
    \end{minipage}
    }
    \caption{Prompt for NER.}
    \label{fig:prompt-ner}
\end{figure*}

\begin{figure*}
    \centering
    \noindent
    \fbox{\small 
    \begin{minipage}{\textwidth}
        \textbf{developer} \\
        Your task is named entity disambiguation. Given a target entity (provided with a triple for context, where the target entity is the object), decide whether it is identical to any of the candidate entities listed below. Return only the letter of the option.\\
        \ \\
        \textbf{user} \\
        Target entity: [target entity]\\
        Context triple: [subject, relation, target entity]\\
        Options:\\
        A. Entity: [candidate entity label]\\
        Description: [candidate entity description]\\
        ...\\
        F. None of above.\\
        G. Unsure - the case is ambiguous/there is not enough information to decide.
    \end{minipage}
    }
    \caption{Prompt for the Triple-Context NED.}
    \label{fig:prompt-ned1}
\end{figure*}

\begin{figure*}[ht]
    \centering
    \noindent
    \fbox{\small 
    \begin{minipage}{\textwidth}
        \textbf{developer} \\
        Generate a one-sentence description of the target entity. You are given a context triple in the form (subject, predicate, object), where the object is the target entity. \\
        \# Instructions\\
        Use the triple to infer relevant information about the entity. Describe the entity based on what is most defining, well-known. \\
        Avoid repeating the information from the triple, unless really essential. \\
        \# Response Format \\
        Return only the sentence: "Description: [one-sentence description of the target entity]"\\
        \ \\
        \textbf{user} \\
        Entity: [target entity] \\
        Triple: [subject, relation, target entity]
    \end{minipage}
    }
    \caption{Prompt for named entity description generation.}
    \label{fig:prompt-nedg}
\end{figure*}

\begin{figure*}
    \centering
    \noindent
    \fbox{\small 
    \begin{minipage}{\textwidth}
        \textbf{developer} \\
        Your task is named entity disambiguation. Given a target entity (provided with a description for context), decide whether it is identical to any of the candidate entities listed below. Return only the letter of the option.\\
        \ \\
        \textbf{user} \\
        Target entity: [target entity]\\
        Target entity description: [target entity description]\\
        Options:\\
        A. Entity: [candidate entity label]\\
        Description: [candidate entity description]\\
        ...\\
        F. None of above.\\
    \end{minipage}
    }
    \caption{Prompt for the Description-Context NED.}
    \label{fig:prompt-ned2}
\end{figure*}

\begin{figure*}[ht]
    \centering
    \noindent
    \fbox{\small 
    \begin{minipage}{\textwidth}
        \textbf{developer} \\
        Given a target predicate (provided with a triple for context), decide whether it refers to any of the candidate predicates listed below. Return only the letter of the option.\\
        \ \\
        \textbf{user} \\
        Target predicate: [target predicate] \\
        Context triple: [subject, target predicate, object] \\
        Options: \\
        A. Predicate: [candidate predicate label] \\
        Description: [candidate predicate description] \\
        ... \\
        F. None of above.
    \end{minipage}
    }
    \caption{Prompt for relation disambiguation.}
    \label{fig:prompt-pd}
\end{figure*}

\begin{figure*}[ht]
    \centering
    \noindent
    \fbox{\small 
    \begin{minipage}{\textwidth}
        \textbf{developer} \\
        Given a predicate that represents a relationship or action between entities, generate a one-sentence description explaining its meaning.\\
        \# Instructions\\
        Focus on describing the relationship, not the entities themselves.\\
        \# Response Format\\
        Begin the description with 'Indicates...' 
        \ \\
        \textbf{user} \\
        Predicate: [target relation] \\
    \end{minipage}
    }
    \caption{Prompt for relation description generation.}
    \label{fig:prompt-pdg}
\end{figure*}

\begin{figure*}[ht]
    \centering
    \noindent
    \fbox{\small 
    \begin{minipage}{\textwidth}
        \textbf{developer} \\
        Given a target class (provided with a triple for context, where the target class is object), decide whether it refers to any of the candidate classes listed below. Return only the letter of the option.\\
        \ \\
        \textbf{user} \\
        Target class: [target class] \\
        Context triple: [subject, instanceOf, target class] \\
        Options: \\
        A. Class: [candidate class label] \\
        Description: [candidate class description] \\
        ... \\
        F. None of above.
    \end{minipage}
    }
    \caption{Prompt for class disambiguation.}
    \label{fig:prompt-cd}
\end{figure*}

\begin{figure*}[ht]
    \centering
    \noindent
    \fbox{\small 
    \begin{minipage}{\textwidth}
        \textbf{developer} \\
        Generate a one-sentence description for a given conceptual class.\\
        \# Response Format\\
        Return only the sentence: "Description: [one-sentence description of the conceptional class]" 
        \ \\
        \textbf{user} \\
        Class: [target class] \\
    \end{minipage}
    }
    \caption{Prompt for class description generation.}
    \label{fig:prompt-cdg}
\end{figure*}

\end{document}